%% file: main.tex
\documentclass[10pt,twocolumn,letterpaper]{article}

\usepackage{cvpr}              %

\input{preamble}

\definecolor{cvprblue}{rgb}{0.21,0.49,0.74}
\usepackage[pagebackref,breaklinks,colorlinks,allcolors=cvprblue]{hyperref}

\def\paperID{4948} %
\def\confName{CVPR}
\def\confYear{2026}

\title{Goldilocks Test Sets for Face Verification}

\author{
  \textbf{Haiyu Wu$^1$\hspace{1mm} Sicong Tian$^2$\hspace{1mm} Aman Bhatta$^1$\hspace{1mm} Jacob Gutierrez$^1$\hspace{1mm}Grace Bezold$^1$} \\ \textbf{Genesis Argueta}$^1$\hspace{1mm} \textbf{Karl Ricanek Jr.}$^3$\hspace{1mm} \textbf{Michael C. King}$^4$\hspace{1mm} \textbf{Kevin W. Bowyer}$^1$\\
  $^1$University of Notre Dame \\
  $^2$Indiana University South Bend \\
  $^3$University of North Carolina Wilmington \\
  $^4$Florida Institute of Technology
}

\begin{document}
\maketitle

\input{sec/0_abstract}
\input{sec/1_introduction}
\input{sec/2_related-work}
\input{sec/3_hadrian-and-eclipse}
\input{sec/4_comparison}
\input{sec/5_ablation-study}
\input{sec/6_conclusion}

{
    \small
    \bibliographystyle{ieeenat_fullname}
    \bibliography{main}
}

\end{document}

%% file: preamble.tex
\usepackage[utf8]{inputenc} %
\usepackage[T1]{fontenc}    %
\usepackage{url}            %
\usepackage{booktabs}       %
\usepackage{amsfonts}       %
\usepackage{nicefrac}       %
\usepackage{microtype}      %
\usepackage{xcolor}         %
\usepackage{booktabs}
\usepackage{colortbl}
\usepackage{graphicx}
\usepackage{subcaption}
\usepackage{relsize}
\usepackage{pifont}
\usepackage{wrapfig}

\newcommand{\cmark}{\ding{51}}%
\newcommand{\xmark}{\ding{55}}%

\newcommand{\red}[1]{{\textbf{\color{red}#1}}}

\newcommand{\blue}[1]{\textbf{\color{blue}#1}}

%% file: sec/0_abstract.tex
\begin{abstract}
Reported face verification accuracy has reached a plateau on current well-known test sets.
As a result, some difficult test sets have been assembled by reducing the image quality or adding artifacts to the image. However, we argue that test sets can be challenging without artificially reducing the image quality because the face recognition (FR) models suffer from correctly recognizing 1) the pairs from the same identity (i.e., genuine pairs) with a large face attribute difference, 2)
the pairs from different identities (i.e., impostor pairs) with a small face attribute difference, and 3) the pairs of similar-looking identities (e.g., twins and relatives). We propose three challenging test sets to reveal important but ignored weaknesses of the existing FR algorithms. To challenge models on variation of facial attributes, we propose Hadrian and Eclipse to address facial hair differences and face exposure differences. The images in both test sets are high-quality and collected in a controlled environment. To challenge FR models on similar-looking persons, we propose ND-Twins, which contains images from a dedicated twins dataset. The LFW test protocol is used to structure the proposed test sets. Moreover, we introduce additional rules to assemble ``Goldilocks\footnote{\url{https://en.wikipedia.org/wiki/Goldilocks_and_the_Three_Bears}}" level test sets, including 1) restricted number of occurrence of hard samples, 2) equal chance evaluation across demographic groups, and 3) constrained identity overlap across validation folds. Quantitatively, without further processing the images, the proposed test sets have on-par or higher difficulties than the existing test sets that add artifacts to the images.
The datasets are available at: \textcolor{blue}{\url{https://github.com/HaiyuWu/SOTA-Face-Recognition-Train-and-Test}}.
\end{abstract}

%% file: sec/1_introduction.tex
\input{figure/teaser}
\section{Introduction}
To evaluate the FR model performance, LFW~\citep{lfw}, CFP-FP~\citep{cfpfp}, CPLFW~\citep{cplfw}, CALFW~\citep{calfw}, AgeDB-30~\citep{agedb-30}, IJB-B~\citep{ijbb}, and IJB-C~\citep{ijbc} are widely used after ArcFace~\citep{arcface}. 
Recently, the model accuracy on these test sets became saturated, so more challenging test sets were assembled~\citep{xqlfw, mlfw, talfw, tinyface, ijbs}. However, none of the referenced new datasets focuses on the natural variation of face attributes, or difficult image pairs in regular resolution. Hence, this paper aims to fill in this void.

Based on the bias and weakness reported in face recognition~\citep{facial-hair-effect, facial-hair-size-effect, brightness-effect, lights}, we propose two test sets $-$ Hadrian and Eclipse $-$ one focusing on the variation of facial hair and the other focusing on the variation of the exposure of the face. On the one hand, both test sets challenge the FR model only by the variation of the natural facial attribute change. On the other hand, the images are from a commercial dataset\footnote{The institution owning MORPH has granted us permission to redistribute the selected data at no cost!}~\citep{morph}, where the images are mugshots collected in a controlled environment. That is, there are no added artifacts involved in reducing the image quality. In fact, the average image quality is higher than typical in-the-wild images.

There exist test sets that focus on similar-looking persons~\citep{sllfw, Doppelver}, but both are too easy (\emph{i.e.}, accuracy $>$ 96\%) to reflect the difficulty of similar-looking persons in real life. The main reason is that both test sets~\citep{sllfw, Doppelver} collect doppelgangers, which is easier than twins or relatives. Considering that there is no identical twins test set formed in the LFW~\cite{lfw} test protocol, we propose ND-Twins that uses selected twins images from a dedicated dataset~\citep{nd-twins}, where the images are collected in a relatively controlled environment but with variations in facial attributes and indoor / outdoor acquisition differences.

To achieve the Goldilocks-level test sets, we restrict the number of occurrences of an image across the 6,000 image pairs to no more than 6, balance demographic groups to the extent possible,  and ensure identity-disjointness across validation folds. As for the quantitative results, we test 15 models (5 algorithms~\citep{arcface, curricularface, magface, adaface, uniface} each trained with 3 different datasets~\citep{arc2face, webface260m, glint360k}) on the existing test sets that follow the LFW test protocol. The average accuracy on Hadrian, Eclipse, and ND-Twins are 1.42\%, 11.23\%, and 22.48\% lower than the most challenging attribute-emphasized test set, CPLFW. Moreover, Eclipse and ND-Twins are more challenging than MLFW~\citep{mlfw} and XQLFW~\citep{xqlfw}, which are intended to largely reduce the face information by adding synthetic face masks and reducing the image quality. Lastly, these three test sets can be useful to evaluate the quality of synthetic datasets~\citep{synface, idiff23, dcface, sface224, vec2face}.
The main contributions of this work are as follows.
\begin{itemize}
\item The proposed test sets reveal that FR models suffer from natural variations of facial attributes and similar-looking tasks, enabling new challenging research directions in face recognition.
\item We introduce additional rules for assembling ``Goldilocks'' level test sets, including restricted hard sample occurrence, equal chance evaluation across demographic groups, and constrained identity overlap across cross-validation folds.
\end{itemize}

%% file: figure/teaser.tex
\begin{figure}[t]
    \centering
        \includegraphics[width=\linewidth]{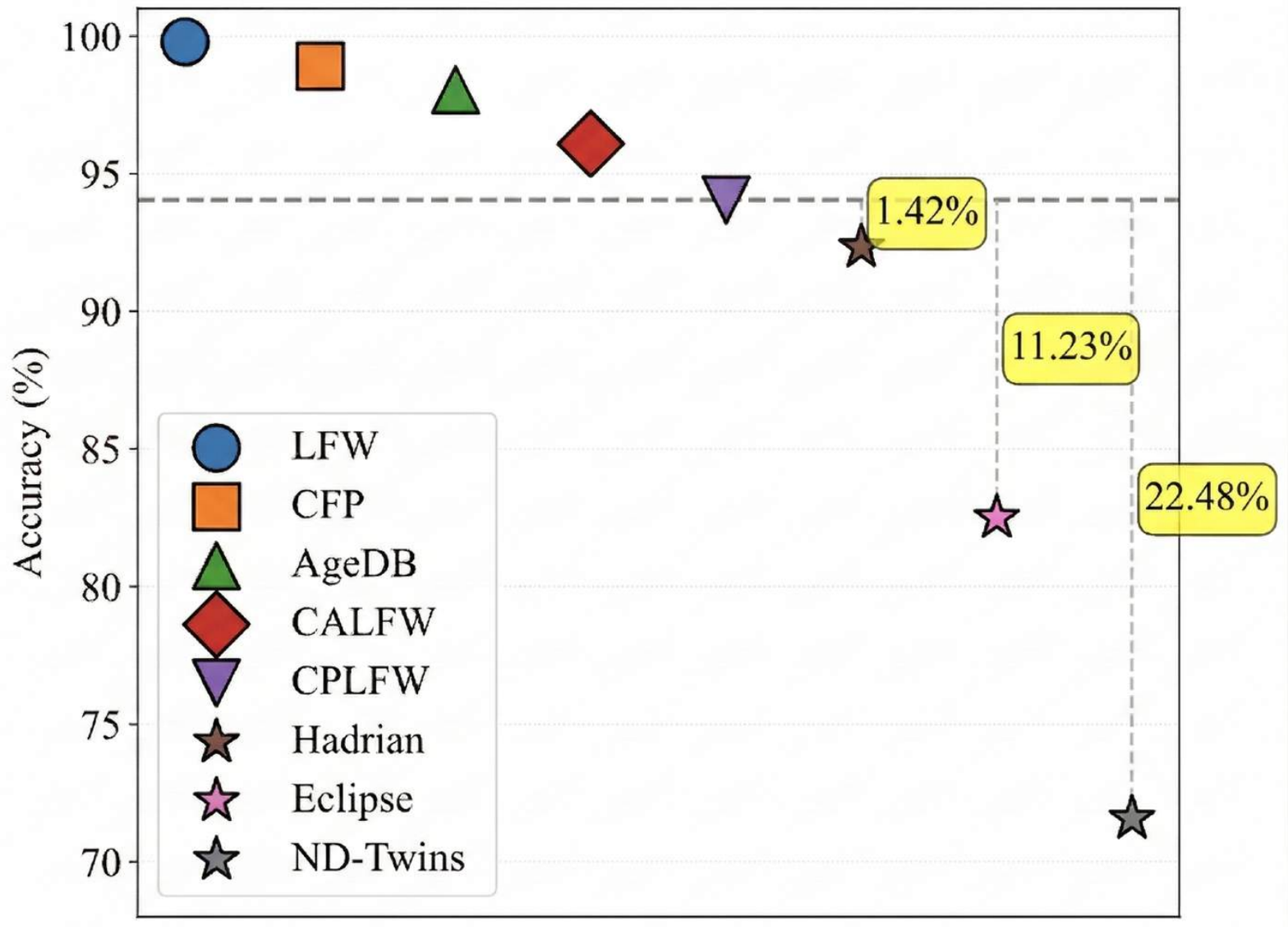}
   \caption{Accuracy comparison among the test sets that have \textbf{natural} faces. The average accuracy reported in Table~\ref{tab:test-performance} is used to indicate the difficulty of the proposed face verification benchmarks (shown in colored $\bigstar$). Moreover, the proposed benchmarks are even harder than those test sets with \textbf{artifacts} (\emph{e.g.,} synthetic face masks and low-resolution) added to face images (see in Figure~\ref{fig:other-test}).
   }
\label{fig:teaser-figure}
\end{figure}

%% file: sec/2_related-work.tex
\section{Literature Review}

\textbf{FR test sets with natural facial attribute variation.}
LFW~\citep{lfw}, YTF~\citep{ytf}, MegaFace~\citep{megaface}, and IJB family~\citep{ijba, ijbb, ijbc} challenge the FR model with mixed attributes, which reflect the performance of the model in the general case. Meanwhile, some test sets are designed to evaluate the model performance in the variation of a single attribute. CFP-FP~\citep{cfpfp} and CPLFW~\citep{cplfw} contain image pairs with a large variation in head pose. AgeDB-30~\citep{agedb-30} and CALFW~\citep{calfw} contain image pairs with a large variation in age. These test sets are widely used, but they yield accuracy that saturates quickly as the scale of the training sets increases, see Table~\ref{tab:test-performance}. 

However, instead of further investigating challenges on natural facial attributes, people start reducing the image quality or adding artifacts to the images. MeGlass~\citep{meglass} emphasizes differences caused by wearing glasses. MLFW~\citep{mlfw} was generated based on LFW for face verification with masks during the COVID-19 pandemic. XQLFW~\citep{xqlfw} compiles challenges for low-quality face recognition. TALFW~\citep{talfw} uses a transferrable adversarial algorithm to disrupt the identity by adding artifacts. All these datasets inject image synthesis into the original images. IJBS~\citep{ijbs}, TinyFace~\citep{tinyface}, and BRAIR~\citep{briar} focus on low-resolution image pair recognition.

To fill this hole in face recognition research, the proposed test sets, Hadrian and Eclipse, focus on facial hair attribute variations and face exposure differences. The images are mugshots, with mostly frontal head pose and neutral facial expression. Moreover, these two test sets are good indicators for evaluating the intra-class variation of the synthetic datasets~\citep{synface, sface22, digiface-1m, dcface, idiff23, vec2face}.

\paragraph{FR test sets with similar-looking pairs.}
Similar-looking tasks (\emph{i.e.}, doppelgangers, twins, relatives) are challenging but lack test sets to reflect such difficulty. SLLFW~\citep{sllfw} and DoppelVer~\citep{Doppelver} contain image pairs of doppelgangers, but existing FR models achieve over 97\% accuracy. This motivates us to assemble an identical (monozygotic) twin test set $-$ ND-Twins, where the same existing FR models achieve 71.57\% accuracy, on average, reflecting the difficulty of discriminating between identical twins. Moreover, there are some other twins related face verification test sets~\cite{twins, vijayan2011twins}, but they are not formed in the LFW test protocol, which hinders the studies in identical twins.

%% file: sec/3_hadrian-and-eclipse.tex
\section{Assembling Goldilocks-level test sets}
The LFW~\citep{lfw} evaluation protocol is a de facto standard for face verification test sets, so our proposed datasets are organized in this way. However, previous test sets are observed to have other potential problems, such as too frequent occurrence of a single image, unevenly distributed demographic groups, and images of the same identity spread across validation folds. Therefore, we applied constraints on these three aspects in assembling our test sets, aiming to assemble Goldilocks-level test sets.

\subsection{Image resource of each proposed test set}
\textbf{Hadrian and Eclipse.} MORPH5~\citep{morph} was assembled from public records. Images were acquired under controlled conditions, including nominally frontal pose, neutral expression, consistent indoor lighting, and a uniform gray background.  Age and demographic meta-data are associated with each image. The two demographics most represented in MORPH are African-American and Caucasian. These features reduce our work for data curation and attribute control.

\input{figure/hadrian-example}
\textbf{ND-Twins.} There are 216 twin pairs from 413 identities in the Twins Challenge dataset~\citep{nd-twins} and each identity has at least 10 images. The face images are high-quality but with the variations on eyeglasses, head pose, and indoor / outdoor environment. The data was collected at the Twins Days festival in Twinsburg, Ohio in August 2009
and August 2010, which ensures the accurate labels of twin pairs.

\subsection{Selection of Facial Hairstyle Pairs for Hadrian}
Facial hair attributes were predicted (using the model from \citep{logicnet}) for 155,683 images from 21,106 African-American males (AAM) and 97,666 images from 10,891 Caucasian males (CM).
The facial hairstyle classification consists of detailed attributes, including beard area, beard length, and mustache. Prior research~\citep{facial-hair-effect, facial-hair-size-effect} has shown that a larger difference in facial hair attributes decreases similarity values and that pairs with mustaches exhibit the highest similarity values. Therefore, to represent challenging facial hairstyle conditions, our target attributes were clean-shaven with no mustache (no-facial-hair) and those with a mustache connected to a full beard (full-facial-hair).  
Genuine pairs for the test set are targeted to be \{no-facial-hair, full-facial-hair\}, and impostor pairs are targeted to be \{full-facial-hair, full-facial-hair\}.
To ensure an accurate selection of images for facial hair attribute pairs, a threshold of 0.9 was applied for the algorithm in \citep{logicnet}, resulting in 21,366 CM and 15,481 AAM images classified as no-facial-hair, and 21,831 CM and 23,221 AAM images classified as full-facial-hair.
We used the aforementioned face recognition model to extract features and calculate the similarity for both genuine and impostor pairs for AAM and CM independently. 
We randomly chose 7,000 genuine and impostor pairs, respectively, for further analysis.
Allowing for the possibility of some identity label noise, impostor pairs with a similarity value higher than 0.7 and genuine pairs with a similarity lower than 0.3 were excluded from selection.

MORPH5 has a range of ages for each identity. To mitigate the impact of age differences, we excluded genuine pairs with an age difference greater than 5 years. Additionally, to prevent any single image from disproportionately influencing the results, we restricted each image to occur in no more than 3 genuine and 3 impostor pairs. These measures reduced the number of genuine pairs to 2,205 for AAM and 1,635 for CM, and impostor pairs to 4,180 for AAM and 1,574 for CM.

We noted that the selected image pairs still exhibited identity label noise and inaccuracies in attribute pairs.
Examples are in Figure~\ref{fig:hadrian-dropped-example}.
Therefore, we performed a manual review to identify and rectify identity label noise, closely matched facial hairstyles in genuine pairs, and significantly different facial hair styles in impostor pairs. As a result, 543 genuine pairs and 279 impostor pairs were eliminated due to incorrect attribute pairing, and 44 pairs were discarded due to identity label noise. All the images are cropped and aligned by img2pose~\citep{img2pose}

\textbf{Hadrian} consists of 1,500 AAM genuine pairs, 1,500 AAM impostor pairs, 1,500 CM genuine pairs, and 1,500 CM impostor pairs. Similar to LFW~\citep{lfw}, CFP-FP~\citep{cfpfp}, CPLFW~\citep{cplfw}, CALFW~\citep{calfw}, and AgeDB-30~\citep{agedb-30}, accuracy is estimated as the mean accuracy from 10-fold cross-validation. In line with the format of previous test sets, each of the 10 folds contains 300 genuine and 300 impostor pairs. We arranged the image pairs by identity to ensure that the genuine pairs of each fold do not share identities with the genuine pairs in other folds, which is an aspect overlooked in earlier datasets. The examples can be found in Figure~\ref{fig:hadrian-example}.

\input{figure/hadrian-dropped-examples}
\subsection{Selection of Under- / Over-exposure Pairs for Eclipse}
Wu et al.~\citep{brightness-effect} categorized MORPH3 into five exposure groups: Strongly Underexposed (SU), Underexposed (U), Middle-exposed (M), Overexposed (O), and Strongly Overexposed (SO), based on percentile boundaries of face region brightness. To create a challenging face exposure test set, the ideal composition for genuine pairs would be (SU,SO), and for impostor pairs, (SO,SO) or (SU,SU). However, applying the same percentile boundaries used by \citep{brightness-effect} for image selection significantly reduces the capacity of the image pool, given that MORPH is a controlled-acquisition dataset.

To eliminate the influence of facial hair, only no-facial-hair images were included in Eclipse. Specifically, it comprises 15,481 African-American male (AAM) images, 20,630 African-American female (AAF) images, 21,366 Caucasian male (CM) images, and 157,025 Caucasian female (CF) images. Following the methodology of~\citep{brightness-effect}, we determined the exposure level for each face image and then selected images from the tails of the exposure distribution for each demographic group. The initial candidate pool was 25\% of AAM images, 20\% of AAF images, 20\% of CM images, and 15\% of CF images from both the upper tail for the high exposure image pool and the lower tail for the low exposure image pool. To identify the most challenging pairs, we used the same face recognition model to extract the features and calculated the similarity of genuine and impostor pairs. Genuine pairs were defined as \{high exposure, low exposure\}, and impostor pairs as either \{high exposure, high exposure\} or \{low exposure, low exposure\}. Using the same threshold values of Hadrian to minimize identity label noise, we selected 2,098 AAF, 1,808 AAM, 7,000 CF, and 3,870 CM pairs for genuine pairs, and 6,000 impostor pairs for each demographic group in both scenarios. After controlling for age differences and image occurrence frequency, the most difficult 1,000 image pairs were chosen for both genuine and impostor pairs across each demographic.
To curate the dataset further, a manual review was conducted to drop identity noise and misclassified %
exposure attribute pairs.
Ultimately, 225 genuine pairs, 184 (high exposure, high exposure) impostor pairs, and 379 (low exposure, low exposure) impostor pairs were excluded from the dataset. No identity noise is detected. The rest of the steps are the same as Hadrian's. 

\input{figure/eclipse-examples}
\textbf{Eclipse} comprises 6,000 image pairs, with each demographic group contributing 750 genuine pairs and 750 impostor pairs. For each demographic, some impostor pairs have over-exposure in both images and some have under-exposure in both images. The examples are in Figure~\ref{fig:eclipse-example}.

\input{figure/nd-twins}
\subsection{Selection of twin pairs for ND-twins}
Unlike the MORPH dataset, the Twins Challenge dataset only has 14,903 images from 413 identities. Initially, we selected the 6,000 image pairs based on the calculated cosine similarity, but none of the FR models can perform better than a random guess (\emph{i.e.}, 50\%). To reduce the difficulty to a proper degree, we randomly select image pairs based on either head pose or indoor / outdoor environment. For genuine pairs, the image pairs have the yaw pose formed as (0$^\circ$, $\pm$45$^\circ$) or the environment formed as (indoor, outdoor). For impostor pairs, the image candidates with a profile pose are dropped from the pool. After this, we selected the most difficult 7,000 impostor pairs and 7,000 genuine pairs as the candidates based on the cosine similarity.

Since the Twins Challenge dataset does not have demographic labels, we use FairFace~\citep{fairface} to obtain the demographic predictions. The statistical result shows that there are 363 White identities and 50 Black identities. Hence, balancing the demographic groups is not practical due to the limitation of the image resource. As for identity disjoint control, one identity occurs in nine folds, and the other identities spread within one fold. The rest of the dataset creation is the same as Hadrian's. Lastly, the proposed ND-Twins contains 6,000 pairs from 6,538 unique images, with 85\% pairs from the White group and 15\% pairs from the Black group. Examples are in Figure~\ref{fig:nd-twins-example}.

Overall, the pipeline to assemble a Goldilocks-level test set consists of 1) reducing the effect of too difficult samples by controlling the image occurrences, 2) balancing the occurrences of demographic groups, and 3) ensuring the identity disjoint between image folds. Due to the difficulty of data collection for twins, the proposed ND-Twins does not perfectly match the second and third conditions. Considering the absence of the test sets focusing on twin pairs, it is still valuable to FR research and we will try to enlarge the scale of the database in future work.

%% file: figure/hadrian-example.tex
\begin{figure*}[t]
    \centering
    \begin{subfigure}[b]{1\linewidth}
    \captionsetup[subfigure]{labelformat=empty}
        \begin{subfigure}[b]{\linewidth}
            \includegraphics[width=\linewidth]{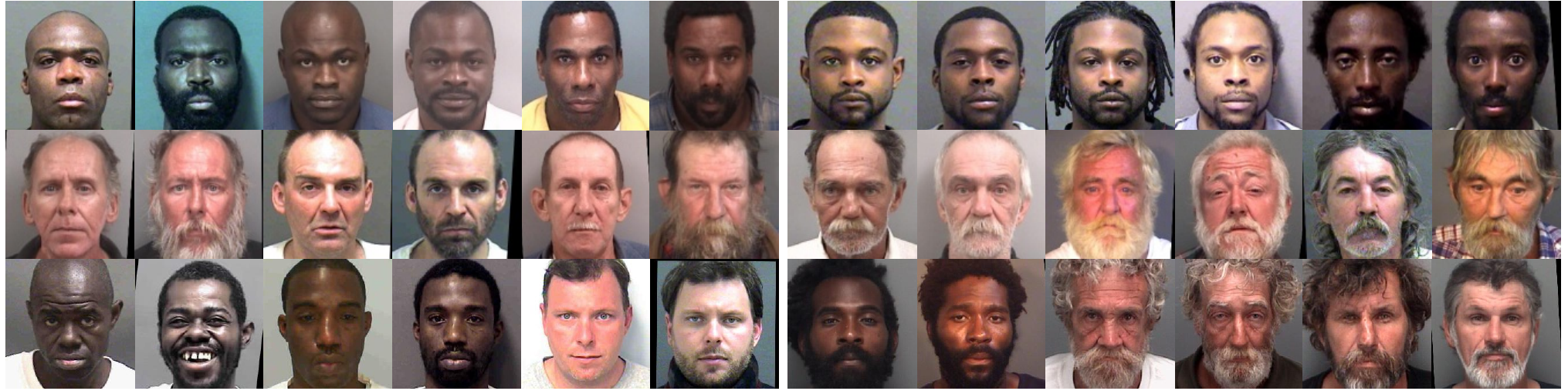}
        \end{subfigure}
    \end{subfigure}
   \caption{Examples of genuine (\textbf{left}) and impostor (\textbf{right}) pairs in Hadrian. The genuine and impostor pairs are formed in \{no-facial-hair, full-facial-beard\} and \{full-facial-hair, full-facial-hair\}.
   }
\label{fig:hadrian-example}
\end{figure*}

%% file: figure/hadrian-dropped-examples.tex
\begin{figure*}[t]
    \centering
    \begin{subfigure}[b]{1\linewidth}
    \captionsetup[subfigure]{labelformat=empty}
        \begin{subfigure}[b]{\linewidth}
            \includegraphics[width=\linewidth]{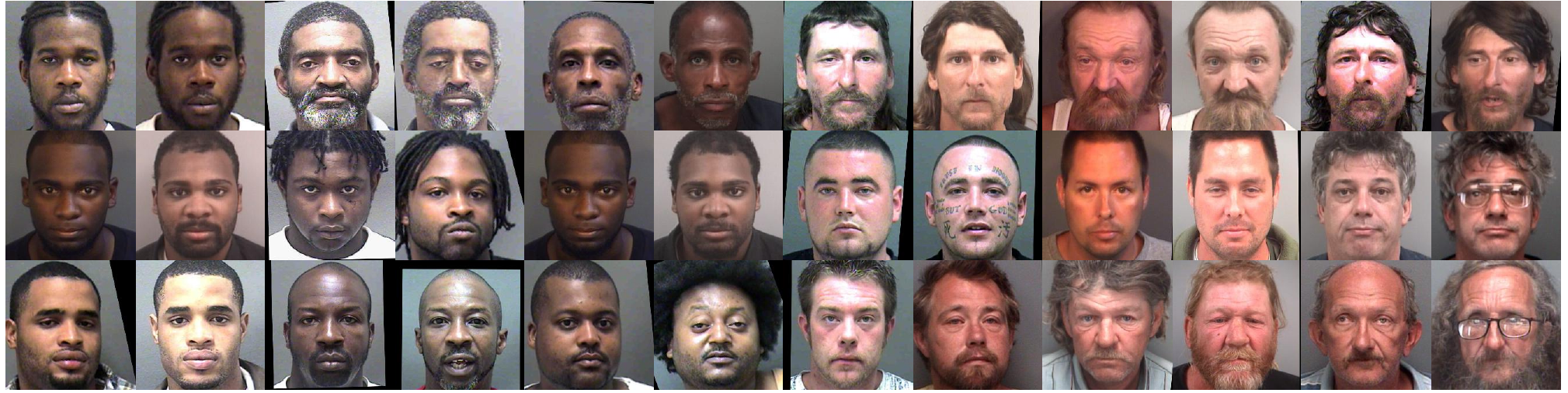}
        \end{subfigure}
    \end{subfigure}
   \caption{Dropped examples during Hadrian assembling. The first row shows the detected pairs of identity noise (\emph{i.e.}, same identity but labeled as different.). The second shows the genuine pairs with unexpected attributes. The third row shows the impostor pairs with unexpected attributes.
   }
\label{fig:hadrian-dropped-example}
\end{figure*}

%% file: figure/eclipse-examples.tex
\begin{figure*}[t]
    \centering
    \begin{subfigure}[b]{1\linewidth}
    \captionsetup[subfigure]{labelformat=empty}
        \begin{subfigure}[b]{\linewidth}
            \includegraphics[width=\linewidth]{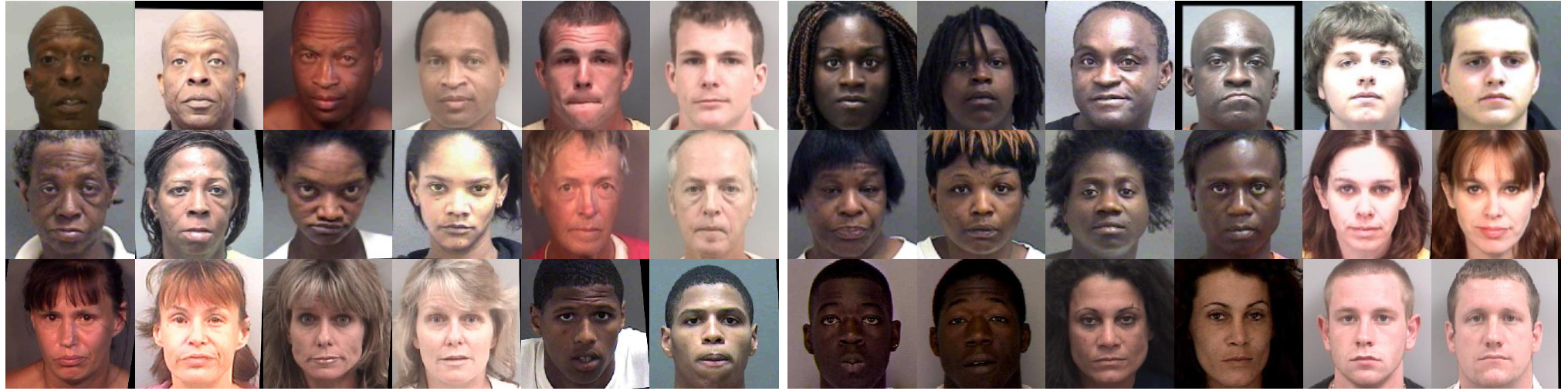}
        \end{subfigure}
    \end{subfigure}
   \caption{Examples of genuine (\textbf{left}) and impostor (\textbf{right}) pairs in Eclipse. The genuine pairs are formed in \{low-exposure, high-exposure\}. The impostor pairs are formed in \{high-exposure, high-exposure\} or \{low-exposure, low-exposure\}.
   }
\label{fig:eclipse-example}
\end{figure*}

%% file: figure/nd-twins.tex
\begin{figure*}[t]
    \centering
    \begin{subfigure}[b]{1\linewidth}
    \captionsetup[subfigure]{labelformat=empty}
        \begin{subfigure}[b]{\linewidth}
            \includegraphics[width=\linewidth]{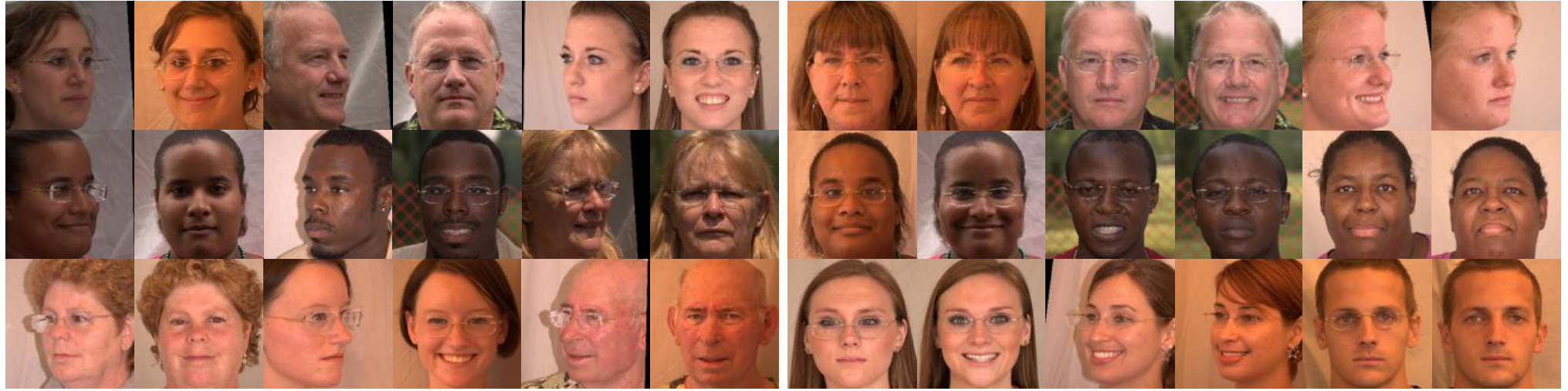}
        \end{subfigure}
    \end{subfigure}
   \caption{Examples of genuine (\textbf{left}) and impostor (\textbf{right}) pairs in ND-Twins. The images in genuine pairs are different in head pose and acquisition environment. The impostor pairs are randomly picked from the image pool with the profile pose image excluded.
   }
\label{fig:nd-twins-example}
\end{figure*}

%% file: sec/4_comparison.tex
\section{Position in the ``Zoo'' of test sets}
Various test sets have been developed to assess the accuracy of face recognition models. 
The two primary evaluation protocols are the 10-fold cross-validation exemplified by LFW~\citep{lfw}, and the True Accept Rate at False Accept Rate (TAR@FAR) exemplified by the IJB family~\citep{ijba,ijbb,ijbc}.  The datasets proposed in this work align with the LFW paradigm, so comparisons with the IJB family are not included.

\input{table/stasts_table}
Table~\ref{table:stats} illustrates the rapid expansion of the LFW test set family to cover diverse challenges such as pose~\citep{cplfw}, age~\citep{calfw}, occlusion~\citep{mlfw}, adversarial attacks~\citep{talfw}, similar-looking~\citep{sllfw}, and image quality~\citep{xqlfw}. In addition to these, AgeDB-30~\citep{agedb-30} and CFP-FP~\citep{cfpfp} are widely recognized for age and pose challenges. A recent dataset, DoppelVer~\citep{Doppelver}, is trying to evaluate models on doppelganger pairs. In contrast, the distinct advantages of the proposed test sets are summarized as follows.

\paragraph{\textbf{Lower repetition of images across pairs.}} The difficulty level of a test set is often influenced by the frequency of challenging examples or noise, which themselves result from a variety of difficult factors. To maintain the integrity and focus of the proposed datasets, we limited the occurrence of each image to no more than 6 times (\emph{i.e.}, no more than 3 times for both genuine and impostor pairs). This largely reduces the evaluation bias arising from hard samples. For example, without this constraint, none of the FR models can achieve better than a random guess (\emph{i.e.}, 50\%) on ND-Twins and less than 55\% on Hadrian and Eclipse.

\paragraph{\textbf{Equal chance of evaluation across demographic groups.}} 
We obtain the estimated race labels by using Fairface~\citep{fairface}. FairFace has two options, for coarse or fine-grained classification. For simplicity, we chose the coarse one, which has White, Black, Indian, and Asian as the race classes. To obtain accurate labels, we apply a majority vote for each identity, where the race label of that identity is decided based on the major predicted race classes. The fraction of each race in LFW, CALFW, CPLFW, CFP-FP, AgeDB-30, and ND-Twins are shown in Figure~\ref{fig:race-gender}. Note that, the MORPH dataset has demographic labels, so Hadrian and Eclipse are not involved in the estimation. \input{figure/race-gender}From the result, the dominance of ``White'' individuals in these datasets implies that models performing best on these datasets may not achieve similar success across other racial groups. Hence, balancing the number of pairs across demographic groups is necessary. Our proposed datasets, Hadrian and Eclipse, ensure an equal number of image pairs for each demographic group, providing a more comprehensive estimate of model accuracy. Due to the difficulty of accurate data collection, ND-Twins does not maintain such a feature. 
\vspace{-3mm}

\input{table/benchmark-performance}
\paragraph{\textbf{More Challenging:}} Table~\ref{tab:test-performance} presents the accuracy values of five face recognition methods, each trained using the original MS1MV2~\citep{arcface}, WebFace4M~\citep{webface260m}, and Glint360K~\citep{glint360k} datasets, across five standard test sets and ours. It is important to note that all models underwent training on the same platform, ensuring consistency in both training and testing procedures, albeit with the default configurations for each method. The results reveal that the average accuracy on CPLFW, the most challenging dataset prior to this study, is 1.42\% higher than on Hadrian, 11.23\% higher than on Eclipse, and 22.48\% higher than on ND-Twins. Given that the MORPH images are recognized as good quality and controlled backgrounds, this underscores the greater challenge posed by the Hadrian and Eclipse test sets, emphasizing specific factors not recognized in previous test sets. 

\input{figure/attribute_comparison}Figure~\ref{fig:other-test} compares the difficulty of similar-looking and low-quality test sets with the proposed datasets. On the one hand, the ND-Twins result suggests that DoppelVer~\citep{Doppelver} and SLLFW~\citep{sllfw} fail to reflect the true difficulty of similar-looking tasks. On the other hand, Hadrian and Eclipse results indicate that natural variation in facial attributes has comparable difficulty with significantly reducing the image quality, showcasing the importance of further investigation on facial attribute variation.

%% file: table/stasts_table.tex
\setlength{\tabcolsep}{4.5mm}
\begin{table*}[t]
\centering
\begin{tabular}{l|c|c|c|c|c|c}
\hline
 & \# of IMs & \# of IDs & Focus & Max$_{occur}$ & Demog$_{balanced}$ & Fold$_{Disj.}$\\ \hline
LFW     & 7,701 & 4,281 & General & 12 & \xmark & \xmark \\ 
CPLFW   & 5,984 & 2,296 & Pose & 13 & \xmark & \xmark\\ 
TALFW   & 7,701 & 4,281 & Attack & 12 & \xmark & \xmark\\  
CALFW   & 7,156 & 2,996 & Age & 11 & \xmark & \xmark\\  
MLFW   & 7,156 & 2,996 & Mask & 11 & \xmark & \xmark\\  
XQLFW   & 7,263 & 3,743 & Image Quality & 8 & \xmark & \xmark\\
SLLFW & 6,091 & 2,810 & Similar-looking & 10 & \xmark & \xmark\\ 
CFP-FP   & 4,366 & 500 & Pose & 24 & \xmark & \xmark\\ 
AgeDB-30 & 5,298 & 388 & Age & 28 & \xmark & \xmark\\ 
DoppelVer & 27,693 & 390 & Doppel\"{g}anger &  82 & \xmark & \xmark\\ \hline \hline
\textbf{ND-Twins} & 6,538 & 219 & Twin & \textbf{6} & \xmark & \cmark\\ 
\textbf{Hadrian} & 7,646 & 2,600 & Facial Hair & \textbf{6} & \cmark & \cmark\\ 
\textbf{Eclipse} & 8,254 & 4,182 & Face Exposure & \textbf{6} & \cmark & \cmark\\ 
 \hline
\end{tabular}
\caption{Statistical information of existing and proposed 10-fold cross-validation test sets. Maximum occurrence of an image (Max$_{occur}$), an equal number of images across demographics (Demog$_{balanced}$), and identity disjoint control across cross-validation folds (Fold$_{Disj.}$) are involved to reflect the feature of the proposed dataset.}
\label{table:stats}
\end{table*}

%% file: figure/race-gender.tex
\begin{figure}
    \centering
    \includegraphics[width=\linewidth]{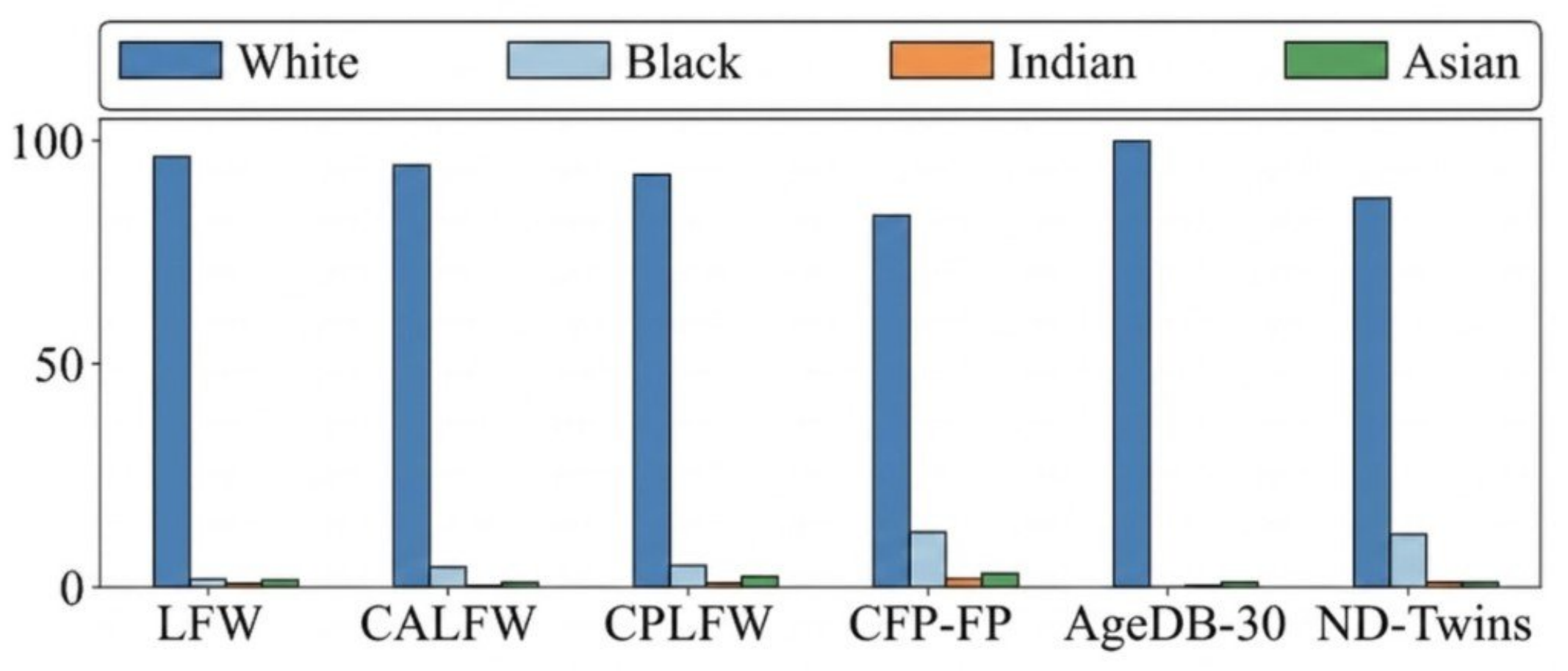}
   \caption{The estimated fraction of each race group in commonly-used test sets, and ND-Twins. The labels are predicted by FairFace~\citep{fairface}. It indicates that the White group dominates the major fraction of these datasets.
   }
\label{fig:race-gender}
\end{figure}

%% file: table/benchmark-performance.tex
\setlength{\tabcolsep}{3.4mm}
\begin{table*}[t]
  \centering
  \begin{tabular}{l|c|c|c|c|c|c|c|c}
    \multicolumn{1}{l}{} & \multicolumn{1}{c}{} & \multicolumn{1}{c}{} & \multicolumn{1}{c}{} & \multicolumn{1}{c}{} & \multicolumn{1}{c}{} & \multicolumn{1}{c}{\includegraphics[width=1cm, height=1cm]{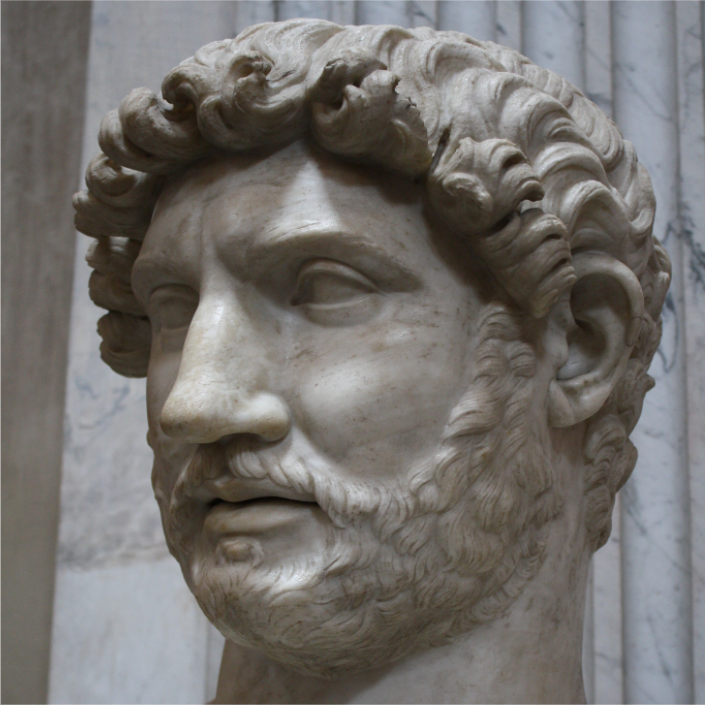}} & \multicolumn{1}{c}{\includegraphics[width=1cm, height=1cm]{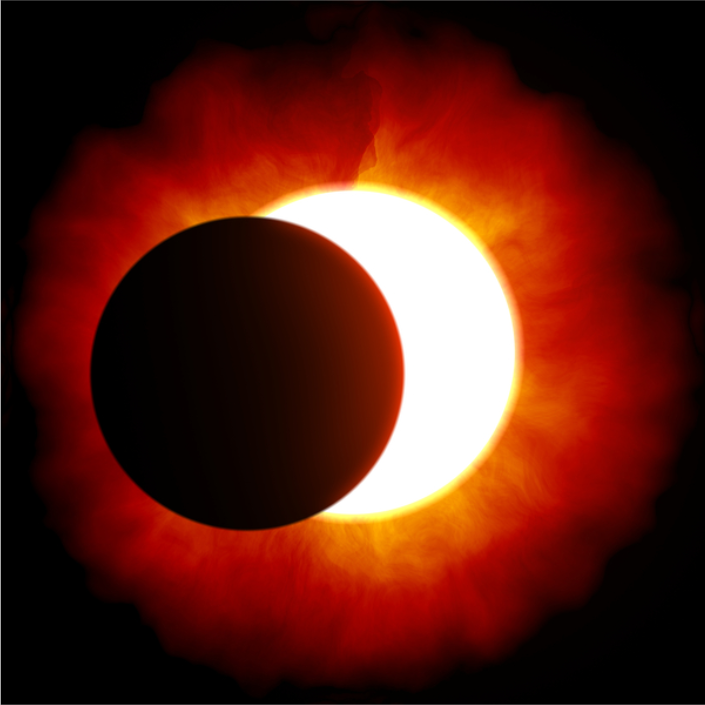}} & \multicolumn{1}{c}{\includegraphics[width=1cm, height=1cm]{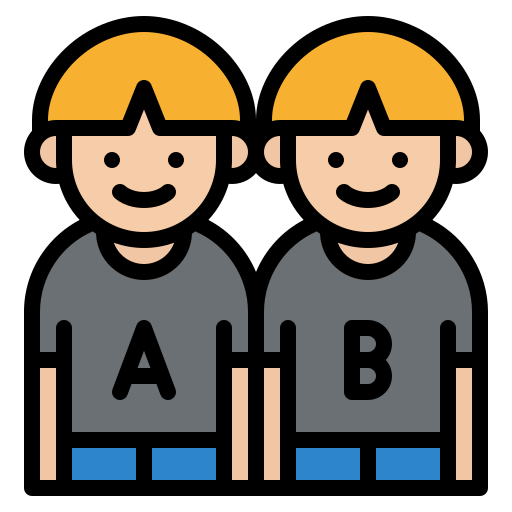}}\\ \hline
    Methods & LFW & CFP & CPLFW & AgeDB & CALFW & \textbf{Hadrian} & \textbf{Eclipse} & \textbf{ND-Twins}\\ 
    \hline
        ArcFace & 99.80 & 98.49 & 93.35 & 98.00 & 96.05 & 91.72 & 81.73 &  68.55 \\ 
        CurricularFace & 99.78 & 98.29 & 92.92 & 98.05 & 96.08 & 91.48 & 82.10 & 66.25 \\ 
        MagFace & 99.82 & 98.40 & 92.95 & 98.22 & 96.02 & 92.80 & 82.67 & 68.58 \\ 
        AdaFace & 99.82 & 98.63 & 93.05 & 98.13 & 96.10 & 92.35 & 82.32 & 68.53 \\ 
        UniFace & 99.78 & 98.49 & 93.28 & 98.02 & 96.10 & 91.63 & 82.18 & 65.73 \\ \hline \hline
        ArcFace & 99.77 & 99.21 & 94.25 & 97.85 & 96.12 & 91.78 & 83.18 & 71.70 \\ 
        CurricularFace & 99.82 & 98.97 & 94.30 & 97.93 & 96.02 & 90.40 & 82.03 & 70.45 \\ 
        MagFace & 99.80 & 99.23 & 94.25 & 97.88 & 95.97 & 90.83 & 82.47 & 71.08 \\ 
        AdaFace & 99.78 & 99.14 & 94.32 & 97.63 & 96.13 & 91.00 & 82.80 & 72.47 \\ 
        UniFace & 99.77 & 99.17 & 94.47 & 97.60 & 96.02 & 90.65 & 82.20 & 71.88 \\ \hline \hline
        ArcFace & 99.82 & 99.07 & 94.68 & 98.30 & 96.15 & 95.68 & 84.12 & 84.67 \\ 
        CurricularFace & 99.78 & 99.11 & 94.70 & 98.40 & 96.17 & 94.63 & 83.50 & 74.27\\ 
        MagFace & 99.82 & 99.14 & 94.47 & 98.20 & 96.15 & 95.22 & 83.83 & 71.93 \\ 
        AdaFace & 99.78 & 99.21 & 94.98 & 98.32 & 96.07 & 95.63 & 83.88 & 75.82 \\ 
        UniFace & 99.73 & 99.09 & 94.58 & 98.22 & 96.18 & 93.53 & 83.18 & 71.53 \\ \hline \hline
        Acc$_{avg}$ & 99.79 & 98.91 & 94.04 & 98.05 & 96.09 & 92.62 & 82.81 & 71.56\\ 
        $\Delta$Acc & \textcolor[HTML]{0a2abc}{\textbf{+5.75}} & \textcolor[HTML]{0a2abc}{\textbf{+4.87}} & 0 & \textcolor[HTML]{0a2abc}{\textbf{+4.01}} & \textcolor[HTML]{0a2abc}{\textbf{+2.05}} & \textcolor[HTML]{E92341}{\textbf{-1.42}} & \textcolor[HTML]{E92341}{\textbf{-11.23}} & \textcolor[HTML]{E92341}{\textbf{-22.48}}\\ \hline
  \end{tabular}
  \caption{Performance of face recognition methods trained with ResNet100 backbone on previous benchmark datasets and ours. Methods are trained on MS1MV2 (\textbf{Top}), WebFace4M (\textbf{Middle}), and Glint360K (\textbf{Bottom}). Acc$_{avg}$ is the average accuracy of each column. $\Delta$Acc is (ours - CPLFW), where CPLFW is the most challenging one among the commonly used test sets.}
  \label{tab:test-performance}
\end{table*}

%% file: figure/attribute_comparison.tex
\begin{figure}
    \centering
    \begin{subfigure}[b]{0.9\linewidth}
        \includegraphics[width=\linewidth]{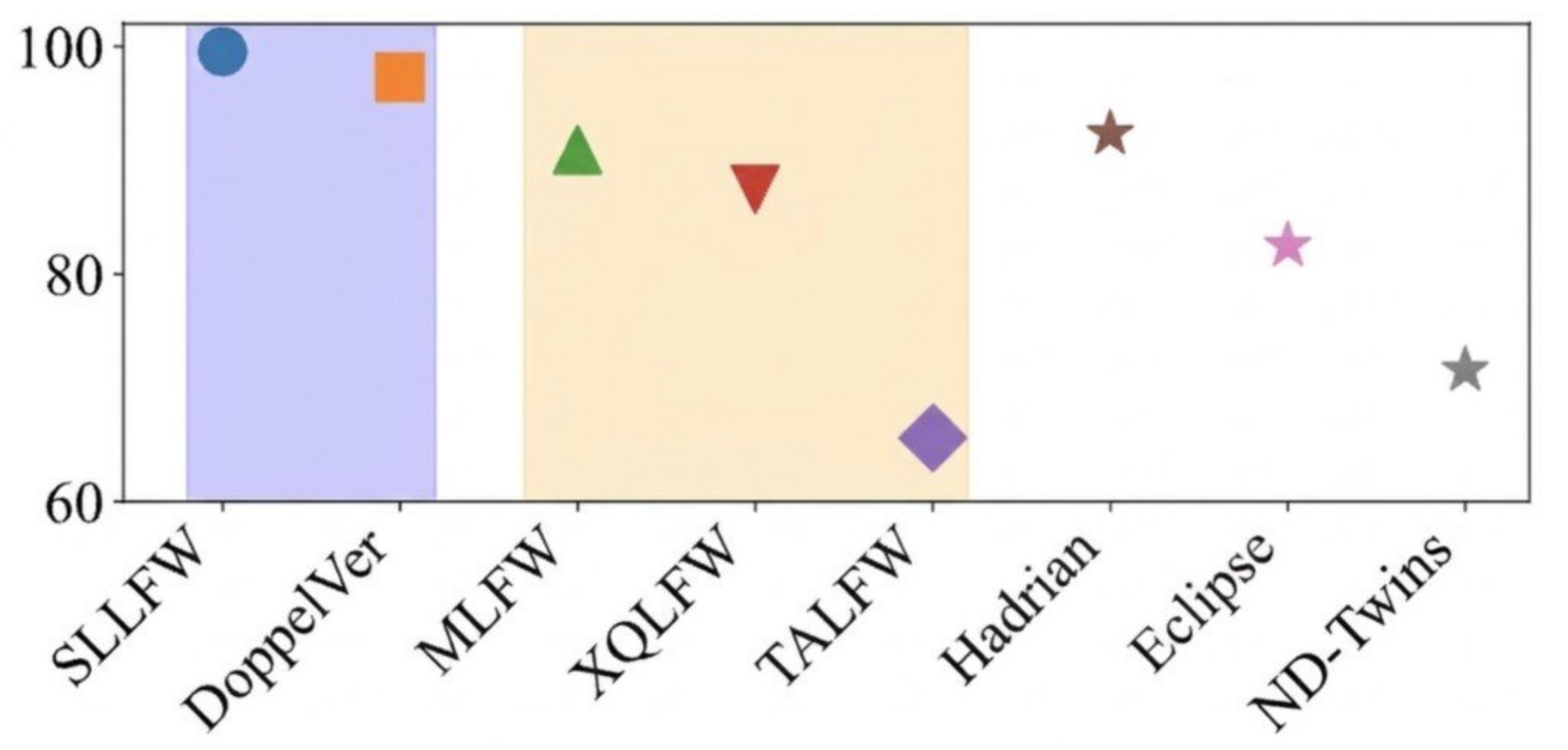}
    \end{subfigure}
    \caption{Comparing with other test sets in average FR accuracy. The results of similar-looking datasets are in \textcolor[HTML]{4A5EF8}{blue}, image-quality-reduction datasets are in \textcolor[HTML]{f5cb84}{orange}, and ours is in the $\star$ shape.}
    \label{fig:other-test}
\end{figure}

%% file: sec/5_ablation-study.tex
\section{Discussion}

\input{table/ablation_study-image-source}

\textbf{The difficulty of Hadrian and Eclipse is not from domain shift.} Current popular training and test sets all contain web-scraped, in-the-wild images, whereas the Eclipse and Hadrian images are from a controlled acquisition scenario. Previous research~\citep{img-video, source-domain} suggests that a change in the data domain can impact face recognition accuracy. Therefore, we conducted an ablation experiment to consider whether the challenging nature of Hadrian and Eclipse could be due simply to domain, rather than the intentional design emphasizing facial hairstyle and under-/over-exposure.

We curated another dataset from MORPH with the same number of identities and images, 750 genuine and 750 impostor pairs from each of Caucasian male, Caucasian female, African-American male, and African-American female, totaling 6,000 pairs.
Identities were randomly chosen for each demographic, without regard to facial hairstyle or under-/over-exposure. 
The final test set follows the LFW structure, where each of the 10 folds contains 300 genuine pairs and 300 impostor pairs, with the four demographic groups evenly distributed among the 300 image pairs. Comparing Table~\ref{table:ablation} and Table~\ref{tab:test-performance} reveals that the randomly-selected MORPH5 test set is the easiest dataset (\emph{i.e.}, accuracy $>$ 99\%) for the fifteen different FR models. Thus, the image source domain does not contribute to the challenging nature of Hadrian and Eclipse.
In fact, the MORPH5 image domain is easier than the existing in-the-wild test sets, as might be expected.

\input{table/ablation_study_id-disjoint}
\textbf{Identity disjoint between folds is necessary.} For the 10-fold cross-validation, each time, 9 folds are regarded as the training set to find the best threshold distance to apply on the 10th fold (test set) in order to calculate the accuracy. This naturally brings up the foundational machine learning question -  
\textit{Does identity overlap across the folds affect the accuracy estimate?} To answer this, we used the same pairs in Hadrian and Eclipse but assigned genuine pairs of the same identity to different folds and kept everything else the same. 
In this case, identities in the test fold have the possibility to occur in the 9 train folds. Table~\ref{table:ablation-overlap} shows that having the test identities in the train sets slightly changes the final evaluation. To preserve the rigor of dataset design, our datasets maintain the ID-disjointness.

\textbf{Having equal evaluation chance across demographic groups is important.} Except for demographic-bias-focused datasets, the research area ignores the equal evaluation chance of demographic groups. This causes the dominance of the White group in the test sets, see Figure~\ref{fig:race-gender}. \input{table/ablation_study_demog} Different from them, the proposed Hadrian and Eclipse balanced the occurrence of pairs from demographic groups. Quantitatively, we report the accuracy of each demographic group in Hadrian and Eclipse in Table~\ref{table:ablation-equal}. It shows that the accuracy difference between demographic groups can be over 30.62\%, which suggests the importance of balancing the evaluation chance across demographic groups to provide a more generalized performance evaluation.

\textbf{Possible cheating on patterns is hard.} We design the pairs with fixed attribute patterns in the proposed test sets. This might raise the concerns about the cheating by a specific design. However, we argue that it is hard to achieve so with the following reasons: In the algorithm design perspective, face recognition models are trained in the multi-class classification manner, so even if the attribute pattern is injected to the algorithm, it will hurt the performance on the other test sets as the model is biased to the attribute pattern. In the dataset design perspective, the used attributes \{full-facial-hair, extreme-exposure, and identical twins\} are rarely appearing in the existing training sets. Hence, collecting / assembling a dataset by using the attribute pattern is difficult.

\textbf{Limitations.} The proposed test sets are assembled with 6,000 image pairs, which is consistent with the commonly-used test sets (see Table~\ref{table:stats}), but it is smaller than the IJB family~\citep{ijba,ijbb,ijbc,ijbs}. This can cause concerns about the generalizability of the evaluation. We have the same concern, so we apply the aforementioned rules to involve more images and demographic groups in the dataset. As for the dataset scale, since the whole MORPH and Twins datasets are not available for free, we have negotiated with the owning institutions to obtain permission to distribute the selected test set images for free. 

\textbf{Safeguard for misuse.} Even though the new test sets will be free for everyone to use, we ask the user to provide the full name, the institution name and sign the agreement to protect the data. The proposed datasets can be downloaded at~\url{https://github.com/HaiyuWu/SOTA-Face-Recognition-Train-and-Test/tree/main?tab=readme-ov-file#hadrian-eclipse-and-nd-twins} in non-commercial purpose. A license agreement is needed for ND Twins due to original IRB conditions.

%% file: table/ablation_study-image-source.tex
\setlength{\tabcolsep}{1.5mm}

\begin{table}
\centering
\begin{tabular}{c|c|c|c|c|c}
\hline
          & Arc.& Curricular.& Mag.& Ada.& Uni.\\ \hline
MS1MV2    & 99.82   & 99.83          & 99.83   & 99.78   & 99.82   \\ 
WebFace4M & 99.83   & 99.80          & 99.82   & 99.83   & 99.82   \\ 
Glint360K & 99.83   & 99.85          & 99.83   & 99.82   & 99.82   \\ \hline
\end{tabular}
\caption{Ablation Study for Domain Effect.  Accuracy for a test set randomly selected from MORPH5 is extremely high.  This indicates the challenging nature of Hadrian and Eclipse is from the designed emphasis on facial attributes. ``Face'' is replaced by ``.'' for current and rest of the tables.}
\label{table:ablation}
\end{table}

%% file: table/ablation_study_id-disjoint.tex
\setlength{\tabcolsep}{0.6mm}

\begin{table}
\centering
\begin{tabular}{c|c|c|c|c|c}
\hline
          & Arc.& Curricular.& Mag.& Ada.& Uni.\\ \hline
Hadrian-Disjoint    &  91.72  &  91.48  &  92.80 & 92.35 & 91.63 \\ 
Hadrian-Overlap    &  91.72  &  91.45  &  92.80  & 92.35 & 91.62 \\ 
$\Delta$Hadrian    & 0 &  \blue{+0.03}  & 0 & 0 & \blue{+0.01} \\ \hline\hline
Eclipse-Disjoint  &  81.73  &  82.10  &  82.67 & 82.32 & 82.18 \\ 
Eclipse-Overlap  &  81.68  &  82.10  &  82.60 & 82.28 & 82.13 \\ 
$\Delta$Eclipse    & \blue{+0.05} & 0 & \blue{+0.07} &  \blue{+0.05} & \blue{+0.05} \\ \hline \hline
ND-Twins-Disjoint &  68.55  &  66.25  &  68.58 & 68.53 & 65.73 \\ 
ND-Twins-Overlap  &  68.70  &  66.27  & 68.58  & 68.52 & 65.82 \\ 
$\Delta$ND-Twins   & \red{-0.15} & \red{-0.02} & 0 & \blue{+0.01} & \red{-0.09}  \\ \hline
\end{tabular}
\caption{Ablation study for the necessity of identity disjoint between folds. $\Delta$ is calculated by (Disjoint - Overlap). The methods are trained with ResNet100 on the MS1MV2 dataset. Despite the incremental difference on accuracy, we still recommend rigorously maintaining the ID disjoint across folds.}
\label{table:ablation-overlap}
\end{table}

%% file: table/ablation_study_demog.tex
\setlength{\tabcolsep}{1.2mm}

\begin{table}
\vspace{-3mm}
\centering
\begin{tabular}{l|c|c|c|c|c}
\hline
          & Arc.& Curricular.& Mag.& Ada.& Uni.\\ \hline
Hadrian-AAM    &  97.17  &  96.43  &  97.73 & 97.10 & 96.87 \\ 
Hadrian-CM    &  86.27  &  86.53   & 87.90 & 87.60 &  86.40 \\ 
\hline\hline
Eclipse-AAM    &  95.00  &  96.07   &  96.73  & 95.93  & 94.47 \\ 
Eclipse-AAF    &  73.33  &  73.73  &  74.07 &  73.53 & 73.87 \\ 
Eclipse-CM    &  93.53  &  93.13   &  93.40  &  93.27 & 92.00 \\ 
Eclipse-CF    &  65.33  &  65.47   &  66.47  &  66.53 &  68.40 \\ \hline
\end{tabular}
\caption{Accuracy on four demographic groups in Hadrian and Eclipse. The methods are trained with ResNet100 on the MS1MV2 dataset. The accuracy difference between demographic groups can be over 30.26\%, showcasing the importance of ensuring equal evaluation chances for each demographic group.}
\label{table:ablation-equal}
\end{table}

%% file: sec/6_conclusion.tex
\section{Conclusions}

We curated three new face verification test sets following the general structure of existing test sets but focused on dimensions of difficulty that are not emphasized in any existing test set.
The Hadrian test set focuses on challenging facial hairstyle pairings,
the Eclipse test set focuses on challenging under- / over-exposure pairings,
and the ND-Twins test set focuses on twin pairings.
To maintain the focus of each dataset, the effects of facial hairstyle are minimized in Eclipse and the effects of under- / over-exposure are minimized in Hadrian.
Curating these test sets from the MORPH and Twins Challenge datasets ensures that they are identity-disjoint with all popular web-scraped training sets. Moreover, controlling the number of occurrences of hard samples and balancing the demographic groups result in Goldilocks-level test sets. For accuracy, the proposed test sets are more challenging than commonly used test sets on 15 FR models. Compared with test sets that contain quality-reduced face images, the proposed test sets achieve comparable or higher difficulty degrees.